\documentclass[10pt,twocolumn,letterpaper]{article}

\usepackage{cvpr}              

\usepackage{graphicx}
\usepackage{amsmath}
\usepackage{amssymb}
\usepackage{booktabs}
\usepackage{kantlipsum} 
\usepackage{dblfloatfix}

\usepackage[pagebackref,breaklinks,colorlinks]{hyperref}

\usepackage[capitalize]{cleveref}
\crefname{section}{Sec.}{Secs.}
\Crefname{section}{Section}{Sections}
\Crefname{table}{Table}{Tables}
\crefname{table}{Tab.}{Tabs.}

\def\cvprPaperID{28} 
\def\confName{EarthVision}
\def\confYear{2022}

\begin{document}

\title{VIDI: A Video Dataset of Incidents}

\author{Duygu Sesver\\
Istanbul Technical University\\
Istanbul, Turkey\\
sesverd19@itu.edu.tr
\and
Alp Eren Gençoğlu \\
Istanbul Technical University\\
Istanbul, Turkey\\
gencoglu17@itu.edu.tr 
\and
Çağrı Emre Yıldız \\
Istanbul Technical University\\
Istanbul, Turkey\\
yildizcag@itu.edu.tr 
\and
Zehra Günindi \\
Istanbul Technical University\\
Istanbul, Turkey\\
gunindi18@itu.edu.tr 
\and
Faeze Habibi \\
University of Tehran\\
Tehran, Iran\\
faeze.habibi@ut.ac.ir 
\and
Ziya Ata Yazıcı \\
Istanbul Technical University\\
Istanbul, Turkey\\
yaziciz21@itu.edu.tr 
\and
Hazım Kemal Ekenel \\
Istanbul Technical University\\
Istanbul, Turkey\\
ekenel@itu.edu.tr
}
\maketitle

\begin{abstract}

Automatic detection of natural disasters and incidents 
has become more important as a tool for fast response. 
There have been many studies to detect incidents using still images and text. 
However, the number of approaches that exploit temporal information is rather limited. One of the main reasons for this is that a diverse video dataset with various incident types does not exist. To address this need, in this paper 
we present a video dataset --- Video Dataset of Incidents, VIDI --- that contains 4,534 video clips corresponding to 43 incident categories. Each incident class has around 100 videos with a duration of ten seconds on average. To increase diversity, the videos have been searched in several languages. 
To assess the performance of the recent state-of-the-art approaches, Vision Transformer and TimeSformer, as well as to explore the contribution of video-based information for incident classification, we performed benchmark experiments on the VIDI and Incidents Dataset. We have shown that the recent methods improve the incident classification accuracy. We have found that employing video data is very beneficial for the task. By using the video data, the top-1 accuracy is increased to 76.56\% from 67.37\%, which was obtained using a single frame. VIDI will be made publicly available. Additional materials can be found at the following link: https://github.com/vididataset/VIDI 

\end{abstract}

\section{Introduction}

Our world faces different natural disasters, such as, fires, earthquakes, flooding. It is crucial to take swift action when these types of disasters happen. The human vision system can understand and recognize the type of disaster with a certain confidence. However, it would be hard to use human power in systems that require fast processing. 
The evaluation by experts is also costly. With the improvement of deep learning, applying computer vision techniques to these challenges has become more popular, less costly, and more effective. 


\begin{figure}[!ht]
  \centering
   \includegraphics[width=1\linewidth]{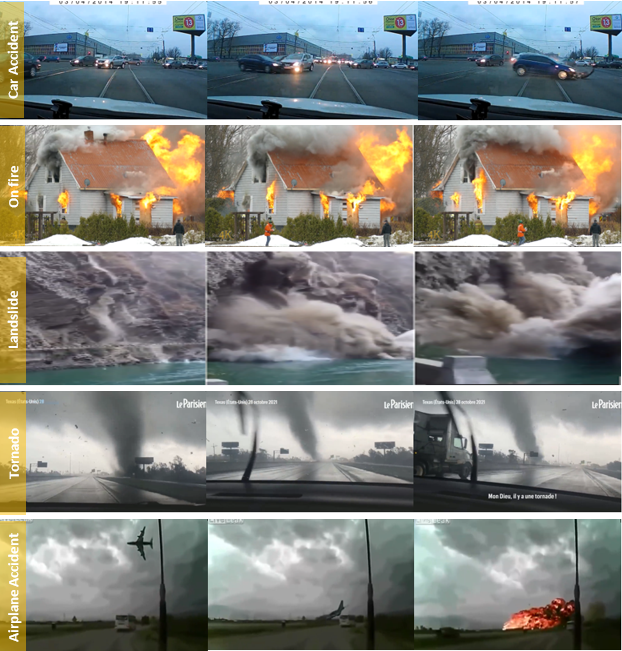}
   \caption{Samples from VIDI which consists of 4,534 video clips of incidents} 
   \label{fig:dataset}
\end{figure}

There have been several studies related to incidents \cite{DBLP:journals/corr/abs-2008-09188, DBLP:journals/corr/abs-1911-09296,DBLP:journals/corr/abs-1809-04094}, and most of them use information based on text and images. To the best of our knowledge, there is no diverse and large video dataset with various incident types. However, temporal information could be useful for the classification of the incidents. 
To enable the researchers to explore the benefits of utilizing temporal information, 
this paper introduces a new video 
dataset for incidents named Video Dataset of Incidents (VIDI). 
The video clips have been collected from YouTube\footnote{https://www.youtube.com/}. Samples from the VIDI can be seen in \cref{fig:dataset}. All video clips have been gathered and labeled by our team. In VIDI, 
when querying YouTube, we used 
different languages and locations to increase the variety. Some video clips are labeled with more than one label, i.e. multiple class labels are allowed in the dataset. To prevent the unrelated information in the videos that could suppress the actions, the clips were preferred to be taken from a single view. Moreover, the clips with sudden frame changes were discarded. Overall, the dataset contains 4,534 video clips labeled with 43 incidents. Each class has around 100 videos. The average video duration is ten seconds. The included incident categories in VIDI are the same as the ones listed in the Incidents Dataset \cite{DBLP:journals/corr/abs-2008-09188}. It additionally contains temporal information and is collected from a different source, that is, while the Incidents Dataset was collected from Flickr and Twitter, VIDI was collected from YouTube. 
Considering these aspects, we believe that VIDI complements the Incidents Dataset.

We run benchmark experiments on the presented dataset using the recent state-of-the-art classification approaches, Vision Transformer (ViT) \cite{dosovitskiy2020image} and TimeSformer\cite{DBLP:journals/corr/abs-2102-05095}. Moreover, we also employed these approaches on the Incidents Dataset. We have shown that these models outperform the ResNet-18 \cite{DBLP:journals/corr/HeZRS15} based approach presented in \cite{DBLP:journals/corr/abs-2008-09188} on the Incidents Dataset. In addition, the obtained results on the VIDI have indicated that the temporal information contributes to the incident classification performance. Using video data, around 9\% absolute increase in classification accuracy has been achieved. 





The remainder of the paper is organized as follows. We review and discuss the related work in Section 2. The Video Dataset of Incidents (VIDI) is introduced in Section 3. Employed classification methods and implementation details are explained in Section 4.
The experimental setup and results are presented and discussed in Section 5. Finally, Section 6 concludes the paper and also provides future research directions.

\section{Related work}
In this part, we will 
review the studies under three different categories: incident detection datasets, methods used to classify incidents and transformer-based deep learning methods. 

\subsection{Incident detection datasets}

Publicly available incident datasets consist of mostly images or text. They 
cover a limited number of classes or have a limited number of examples. Only, recently released Incidents Dataset \cite{DBLP:journals/corr/abs-2008-09188} provides 446,684 images with 43 disaster categories and 49 place categories. 
This dataset consists of images, therefore, it is not possible to build video-based incident classification approaches using it. Another incident dataset with images is xBD \cite{DBLP:journals/corr/abs-1911-09296}, which contains satellite images of six different types of disasters ---volcanic eruption, hurricane, earthquake, flood, tsunami, fire--- from 19 different disaster sites. Even though xBD has 22,068 images, it has a limited number of disaster categories. The last image-based incident dataset that we have examined is ME17-DIRSM \cite{8398414}. This dataset consists of 6,600 images for flooding incidents as positive and negative. That is, this dataset is collected only for a specific incident type and does not contain other types of disasters.

There is a large-scale annotated video dataset 
named FIVR-200K \cite{DBLP:journals/corr/abs-1809-04094}, which focuses on fine-grained video search. In this study, the authors crawled all events from Wikipedia\footnote{https://en.wikipedia.org/wiki/Main\_Page} within the time interval between 2013 and 2017 and filtered the events with two main labels: "armed conflict \& attack" and "disaster \& accident". The video set was created with the YouTube videos, which were queried with the filtered events. FIVR dataset has 225,960 clips, which are labeled with event headlines and names of the videos.

A summary of some existing incident datasets is listed
in \cref{tab:statistics}. For a comprehensive list of incident datasets please refer to \cite{DBLP:journals/corr/abs-2008-09188}. Similar to the Incidents Dataset \cite{DBLP:journals/corr/abs-2008-09188}, we aim to provide a diverse dataset from various types of incidents. Therefore, we utilize the same incident categories that were covered in \cite{DBLP:journals/corr/abs-2008-09188}. As VIDI consists of video clips of the incidents, researchers can benefit from it to develop and assess their incident detection and/or classification systems that incorporate temporal information. 


\begin{table}[!t]
  \centering
  \begin{tabular}{@{}lrrc@{}}
    \toprule
    Dataset Name & \# of data & \# of class  & Type\\
    \midrule
    Incidents Dataset & 446,684 & 43 & Image\\ 
    xBD & 22,068 & 6 & Image \\ 
    ME17-DIRSM & 6,600 & 2 & Image \\
    FIVR &  225,960 & 2 & Video \\ 
    VIDI (Ours) & 4,534 & 43 & Video \\ 
    \bottomrule
  \end{tabular}
  \caption{List of different incident datasets. For a comprehensive list please refer to \cite{DBLP:journals/corr/abs-2008-09188}.
  }
  \label{tab:statistics}
\end{table}

\subsection{Methods used to classify incidents}


Most of the incident classification systems employ deep learning-based approaches to existing social media \cite{58,15,16,54,55,3,45,61,1} and satellite datasets \cite{Gueguen2015LargescaleDD,8398414,47,64}.


In one of the studies to detect fires, a dataset was created with publicly shared 114,098 images gathered from Flickr~\cite{16}. They used multiple feature detectors --– SIFT \cite{sift}, SURF \cite{surf}, Color-SIFT \cite{colorsift}, Color-SURF \cite{colorsurf} –-- to extract the keypoints of the images and the descriptors at the keypoints. By using the k-means algorithm\cite{Jin2010}, the descriptors are clustered to create the visual vocabulary. 
Afterwards, a support vector machine (SVM) \cite{Cristianini2008} classifier is trained and the images are classified with this model. 


In another study \cite{54}, the authors employed three image classification approaches to assess the severity of the incident damage. The used dataset was collected from social media and it contains 210,000 images from five sources. In the first method, they extracted the features by using Pyramidal Histogram of Visual Words (PHOW) and Bag-of-Visual-Words (BoVW)~\cite{bbow} methods, and trained a linear SVM. The second method uses a pre-trained VGG-16 model \cite{Simonyan2015VeryDC} as a feature extractor. As the final approach, they fine-tuned a pre-trained VGG-16 model to classify the damage severities into three classes: None, mild, severe. They reached the best results on all datasets with the fine-tuned VGG-16 model.

Nogueira \etal proposed a convolutional neural network (CNN) based approach to detect flooding from satellite images \cite{8398414}. They used ME17-DIRSM dataset which has a total of 6,600 images for flooding incidents as positive and negative. They used two types of models for training. 
The first one uses dilated convolutions, which do not change the resolution in the learning process. The second one uses deconvolution networks, the encoder part of the network learns to extract the visual features, whereas the decoder part learns to upsample these features. By concatenating the outputs of these models, 
they trained an SVM classifier.

Finally, in \cite{DBLP:journals/corr/abs-2008-09188}, they applied a ResNet-18~\cite{DBLP:journals/corr/HeZRS15} model as a backbone and train the model to jointly recognize the incident type and place categories by using the class-negative loss. 

\section{VIDI}

In this section, we will take a closer look at the created dataset. We will describe how we have collected the videos and checked their quality, as well as we will share the dataset statistics.

\subsection{Overview}

The Video Dataset of Incidents (VIDI)\footnote{https://vididataset.github.io/VIDI/} consists of videos containing natural disasters such as earthquakes, floods, landslides, vehicle accidents such as car, truck, and motorcycle crashes, and the results of these events whether they are damaged, burned, and/or collapsed. The same 43 
incident categories as listed in \cite{DBLP:journals/corr/abs-2008-09188} are included. 
In~\cite{DBLP:journals/corr/abs-2008-09188}, the images were also labeled with 49 different places for diversity. Although we did not label the places, 
we aimed to collect the videos from different locations to provide a similar diversity in our dataset. 

Another approach to increase diversity is searching videos in various languages to get the styles of different cultures and include the region-specific incident events. For example, the appearance of buildings may differ from each other depending on culture, location, and material availability. Wood is commonly used as a building material in America, whereas limestone is more common in the Middle East. Additionally, the precautions and consequences of different climatic events are vastly different in countries where they are expected and those in which they are not.

Mainly, six different common languages have been used in video queries. These languages are English, Turkish, French, Spanish, Simplified Chinese, and Standard Arabic. Apart from these languages, when the amount of collected videos is not sufficient, to collect more videos, Hindi, German, 
and several other languages have also been used in the dataset. The language statistics of the collected video clips are shown in \cref{tab:lang}. However, it may not be possible for these languages to be equally distributed for each class due to the occurrences of specific disasters in some regions. For example, a derecho is an event that exists mostly in North America. Thus, the derecho video clips were mainly collected in English. To overcome this point, we found the list of countries where this disaster occurs and searched in different languages according to these regions. 
\begin{table}[!ht]
  \centering
  \begin{tabular}{@{}lr@{}}
    \toprule
    Language & \# of clip \\
    \midrule
    English & 2140 \\ 
    Spanish & 560 \\ 
    Turkish & 452 \\  
    French & 376 \\  
    Simplified Chinese & 343 \\  
    Standard Arabic & 280 \\ 
    Other & 144 \\
    \bottomrule
  \end{tabular}
  \caption{The number of video clips from different languages in VIDI. We used different languages to increase diversity.  }
  \label{tab:lang}
\end{table}

Another issue that we encountered during the data collection was the translation of the incident keywords between languages. For some of the languages, we had to use neural machine translators. In some cases, the translation of the corresponding incident was not meaning the same in the respective country. The tropical cyclone could be given as an example. When the label "tropical cyclone" is translated to the Arabic language, it is seen that translated word has the same meaning as a tornado. These problems were solved by communicating with native speakers when possible.

Besides the translation issues, we had difficulty in finding videos for some labels, such as nuclear explosion. The existing videos for nuclear explosions were recorded between 1945 to 1963 in the US and USSR. Since atmospheric testing was prohibited after that year, there were a limited number of unique videos with this label. Therefore, the gathered videos mostly contain near duplicates in the positive examples: the same scene has been included multiple times in the dataset from different channels. 

Finally, we allowed multiple labels in our dataset. Some videos show multiple incident types, for example, a burned house is also labeled with the damaged label and a car crash is labeled with both car accident and blocked. 

\begin{figure*}[t]
  \centering
   \includegraphics[width=1\linewidth]{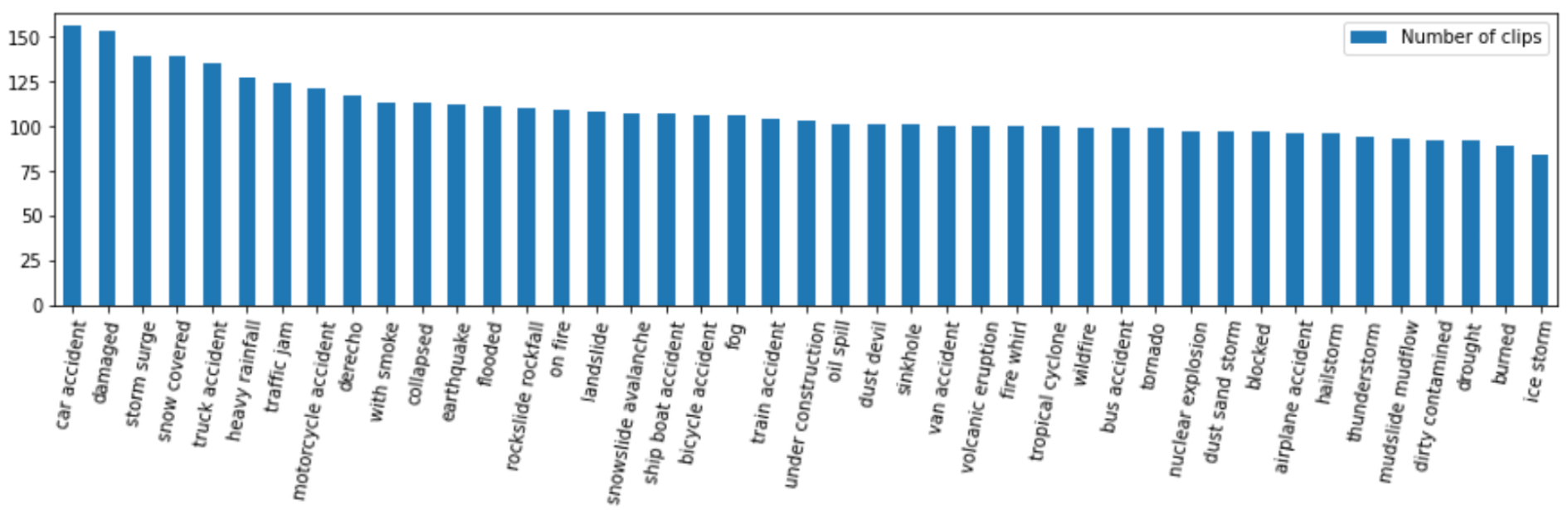}
   \caption{The number of video clips for each class in VIDI. Each class has around 100 video clips.}
   \label{fig:rsm}
\end{figure*}

\begin{table*}[!b]
  \centering
  \begin{tabular}{@{}lccccc@{}}
    \toprule
     Experiments & Optimizer & Loss fn & Base Learning Rate & LR shrinkage & \# of frames \\
    \midrule
    ViT on Incidents Dataset & SGD & CE Loss & 0.001 & lambdaLR & 1\\ 
    TimeSformer on Incidents Dataset & Adam & CE Loss & 0.005 & - & 1\\
    TimeSformer on VIDI & SGD & BCE with Logits & 0.01 & MultiStepLR & 8 \\  
    TimeSformer on VIDI & SGD & BCE with Logits & 0.01 & MultiStepLR & 1 \\ 
    ViT on VIDI & SGD & BCE with Logits & 0.25 & ReduceLROnPlateau & 1\\  
    \bottomrule
  \end{tabular}
  \caption{Training parameters used for training ViT and TimeSformer models on Incidents and VIDI datasets. CE: cross-entropy, BCE: binary cross-entropy, and LR: learning rate.}
  \label{tab:impdet}
\end{table*}

\subsection{Annotation and quality assurance}

All videos have been obtained manually from YouTube by our team members. While querying, the incident labels and locations were combined. For example, we combined the word "fire" with different places for the "on fire" label and queried on YouTube for words similar to "fire in the building", "fire in the forest", etc. Another example is the car accident label. We have searched for car accidents that occurred at different places, such as fields, forests, seasides, and streets. 
For the suitable clips, video ids, starting and ending seconds of the related action, labels, and the language of the video were exported. If possible, all stages of the actions as pre-action/action/post-action were included. Another considered point was that the videos were taken only from a continuous camera view. There are no scene changes in the clips, and almost all of the videos have a one-point perspective. However, we allowed both unsteady and steady shots in the clips.

The video clips which have higher than 720p resolution were downsampled to 720p and downloaded. On the other hand, the videos that have lower resolution were downloaded with their native resolutions. After double-checking the labeled clips, the most common issues we have found were labeling the clip with the wrong label or labeling the clip with a single label, when it contains multiple incidents, such as car accidents under heavy fog. Some labels can be seen as similar to each other, therefore, YouTube users may upload videos with wrong titles, for example, confusing the derecho and tornado events. Furthermore, the users may upload videos that do not include any action associated with the label within the clip duration. By way of illustration, a video with windy environment may have been shot in a tropical area, but the clip does not contain any information related to the region. In that case, we changed the label from "tropical cyclone" to "derecho". We revised the entire collected data for these kinds of situations. 

\subsection{Statistics}

VIDI contains 4,534 unique videos in total. Since some of the videos have more than one label, 
there are 4,767 labels in the dataset. 
Each incident class has around 100 
video clips. The number of videos per class is shown in \cref{fig:rsm}.

We kept the video duration between 2 to 60 seconds. The average video duration is 
ten seconds. The short clips
were chosen from news channels' broadcasts as they provide various disasters, 
however, they generally change the clips in less than two seconds. Therefore, we had to collect the longer clips mostly from documentaries and videos captured by mobile phone cameras.

We left roughly ten clips from each class for testing and validation. Thus, the dataset has 405 and 408 videos in the test and validation sets, respectively. Taking into account the multiple labels, the test set and the validation set contain 434 and 438 labels, respectively. The rest of the set was used for training.

\section{Method}
 
In this section, we will explain the methods and 
provide implementation details. 

\subsection{Architectures}
As the Vision Transformer and TimeSformer are recent state-of-the-art methods for image and video classification, respectively, 
we decided to use these two architectures in our benchmarks. The ViT model was designed for images, whereas the TimeSformer model was designed for videos. Both architectures were pre-trained on the ImageNet dataset and fine-tuned on the Incidents Dataset and VIDI. Vision Transformer architecture 
does not use any convolutional network layers and is based on the transformer encoder blocks. TimeSformer does not contain convolutional network layers, either, and it consists of a self-attention mechanism. Similar to Vision Transformer, TimeSformer also adapts transformer architecture for computer vision tasks and video processing.



\subsection{Implementation details}

We used the PyTorch\footnote{https://pytorch.org/} framework for the implementation. We trained both datasets using two different architectures and performed five different experiments. For the multi-frame fine-tuning process of TimeSformer, eight frames are sampled sequentially from each video in VIDI. For the single-frame fine-tuning process of both architectures, the frame in the middle of the video was selected and given to the model as an input. To fine-tune the TimeSformer model on the Incidents Dataset, each image was assumed as a video that includes only one frame. 

As the pre-trained model, B\_16\_imagenet1k\footnote{https://bit.ly/3CMzSkb} which was trained on ImageNet-21k \cite{imagenet} and fine-tuned on ImageNet-1k datasets were used for Vision Transformer. The divided space-time model that was trained on the K400\footnote{https://bit.ly/3I9TgZi} \cite{DBLP:journals/corr/CarreiraZ17} and K600\footnote{https://bit.ly/36iq1Xi} \cite{DBLP:journals/corr/abs-1808-01340} datasets were used for TimeSformer.

\begin{figure*}[!b]
  \centering
   \includegraphics[width=1\linewidth]{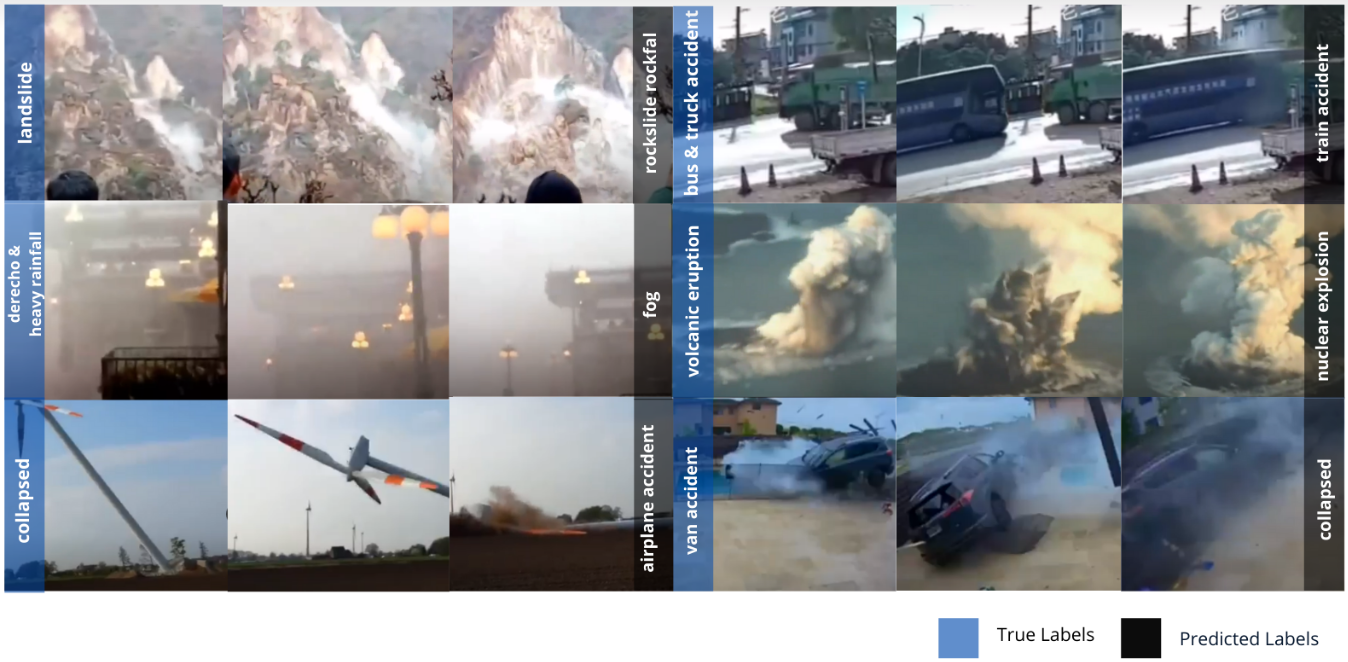}

   \caption{Sample false classifications from VIDI. 
   Three frames are shown from each video, along with the correct labels (on the left), and the predicted labels (on the right). 
   }
   \label{fig:errr}
\end{figure*}

The training parameters used in the experiments are listed in \cref{tab:impdet}. Based on the hyperparameter optimization results on the validation set, 
Adam optimizer \cite{adam} was used for fine-tuning the TimeSformer model on the Incidents Dataset, and the stochastic gradient descent optimization method was used in the rest of the experiments. The cross-entropy loss was preferred on Incidents Dataset, whereas binary-cross entropy with logits loss was chosen for the VIDI. 
Different learning rate schedulers were utilized 
for training. 
Specifically, Lambda LR, ReduceLROnPlateau, and multi-step LR methods were used in the training of ViT and TimeSformer models. Lambda LR method updates the learning rate according to the current epoch number. In our training, the learning rate is reduced in every three epochs. In multi-step LR method, the learning rate is decreased in every five epochs. Finally, in the ReduceLROnPlateau method, the learning rate is reduced if loss change is less than a certain threshold during two epochs. 

\section{Experiments}

We run experiments on both Incidents Dataset and VIDI. Incidents Dataset was already split into training, validation, and test sets. 
While there are 312,742 samples in the training set, there are 17,357 samples in the validation set and 17,255 samples in the test set. VIDI contains 3721 videos for training,  
408 videos for validation, and 405 videos for testing. 
Please note that both datasets have the same incident categories and can include multiple labels per image/video, that is, an image or video might belong to more than one incident category.

Two architectures were trained using only positive incident samples for both datasets to perform incident classification. 
In Timesformer, the prediction was done three times in the testing phase with different crops. For each prediction of a clip, we used the sum method to combine the scores from each crop. Top-1 and top-5 accuracies were used as the evaluation metric.

Multiple experiments were conducted by using both datasets and architectures. One of our 
motivations while collecting VIDI was to explore the benefit of using video data instead of single frame for incident classification. 
To investigate this, we fine-tuned the TimeSformer architecture on VIDI both with a multi-frame and a single frame setup. Multi-frame TimeSformer achieved 76.56\% accuracy, whereas single frame version obtained 67.37\% accuracy. This indicates that, when available, utilizing video information helps to improve incident classification performance. 


In the experiments, we also assessed the performance of the recent state-of-the-art architectures, ViT and TimeSformer, for incident classification in comparison to the proposed approach in \cite{DBLP:journals/corr/abs-2008-09188}, which is based on ResNet-18. 
As can be seen in \cref{tab:res}, ViT and TimeSformer achieved higher accuracies than the published results in \cite{DBLP:journals/corr/abs-2008-09188}. While Resnet-18 based model \cite{DBLP:journals/corr/abs-2008-09188} reaches 77.3\% accuracy, ViT and TimeSformer achieved 78.5\% and 81.47\% top-1 accuracies, respectively, on the Incidents Dataset. 

\begin{table}[!ht]
  \centering
  \begin{tabular}{@{}lc@{}r@{}r@{}}
    \toprule
     Architecture & Dataset & \ Top-1 Acc & \ \ Top-5 Acc \\
    \midrule
    ResNet-18\textsuperscript{\textdagger} & Incidents Dataset & 77.3 &  95.9 \\ 
    ViT & Incidents Dataset & 78.5 & 96.33  \\  
    TimeSformer & Incidents Dataset & \textbf{81.47} & \textbf{96.95}  \\ 
    ViT & VIDI & 61.78  & 86.78 \\
    TimeSformer & VIDI & 67.37 & 90.59  \\
    TimeSformer & VIDI* & \textbf{76.56}  & \textbf{96.51} \\
    \bottomrule
  \end{tabular}
  \caption{
  Top-1 and Top-5 accuracies (\%) of different models on the datasets. 
  *Using multiple frames. In the rest of the experiments, a single frame is used. \textsuperscript{\textdagger}Result was published in \cite{DBLP:journals/corr/abs-2008-09188}.}
  \label{tab:res}
\end{table}

Additionally, we compared the performance of ViT and TimeSformer on both datasets using a single frame as the input. TimeSformer is found to be superior to the ViT on both datasets. The results obtained on the VIDI are lower than the ones obtained on the Incidents Dataset. There could be two main reasons for this: (1) Incidents Dataset contains more samples for training, therefore, the systems were able to learn better models. (2) VIDI contains more difficult samples for classification. 

\begin{figure*}[!ht]
  \centering
    \includegraphics[width=0.8\linewidth]{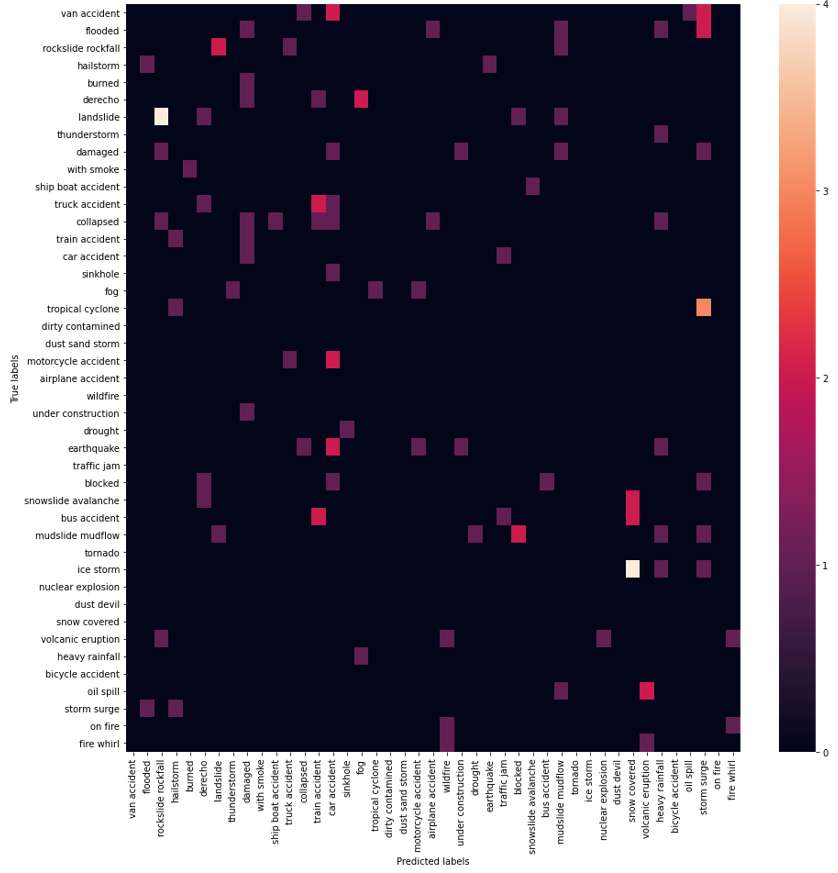}
   \caption{False Prediction/Confusion Matrix 
   }
   \label{fig:confm}
\end{figure*}

We further analyzed the wrong predictions on the VIDI test set. Some sample frames from wrong predictions can be seen in \cref{fig:errr}. As can be observed, in some cases, 
the inputs are visually very similar to the samples of the predicted classes. 
In \cref{fig:confm}, we showed the false prediction matrix. From this matrix, one can observe which classes are mostly confused by the model. 
The most misclassified classes are found to be "ice storm \& snow-covered", "flooded \& storm surge", and "landslide \& rockslide rockfall". 

\begin{figure*}[!ht]
  \centering
    \includegraphics[width=0.8\linewidth]{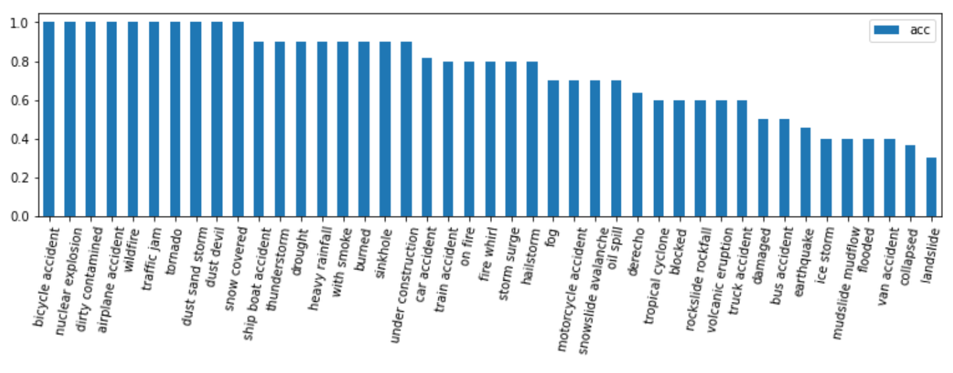}
   \caption{Top-1 accuracies per class 
   }
   \label{fig:acc}
\end{figure*}

Finally, we have calculated the prediction accuracy per class as shown in \cref{fig:acc}. The video clips labeled with "bicycle accident", "nuclear explosion", "dirty contamined", "airplane accident", "wildfire", "traffic jam", "tornado", "dust sand storm", "dust devil", and "snow-covered" have been predicted with higher accuracies than the other classes. It was observed that the lowest accuracy belongs to the "landslide". According to the results, the classes with lower accuracies are mostly mixed within different vehicle accidents. 




\section{Conclusion}

In this paper, we presented a video dataset of incidents, named VIDI. VIDI 
contains 43 different incident categories, for example, earthquake, wildfire, landslide, tornado, ice storm, car accident, nuclear explosion, and  has 4,534 human-labeled incident video clips collected from YouTube. The dataset contains the same incident categories as presented in the Incidents Dataset \cite{DBLP:journals/corr/abs-2008-09188}. As it additionally contains temporal information as well as collected from a different source, i.e. Flickr and Twitter vs. YouTube, we believe that the proposed dataset complements the Incidents Dataset well. 

We run recent state-of-the-art classification models, namely Vision Transformer and TimeSformer, both on VIDI and the Incidents Dataset to provide benchmark results. We observe that the employed approaches perform better than the ResNet-18 based method proposed in \cite{DBLP:journals/corr/abs-2008-09188} on the Incidents Dataset. We also show that utilizing video data improves the incident classification accuracy. 

We will continue to work on the proposed dataset and expand it with class-negative samples. When class-negative samples are included, besides performing incident classification, we will also address the 
incident detection problem 
with an aim to have a robust incident detection system that is suitable to be used in real-time scenarios. 

Lastly, we have only used visual information. 
Yet, the audio information could also be included as supportive  data  for  the  cases  in  which  the incidents  have  similar  environmental  characteristics. 
Thus,  the  audio  could  be  a  meaningful complementary input to be used to identify the correct situation. 

\section*{Acknowledgement}

We would like to thank Ferda Ofli for his support and Ethan Weber for sharing test set of the Incidents Dataset. This study was funded by the Scientific and Technological Research Council of Turkey (TUBITAK) ARDEB 1001 Grant No 121E408 and the Istanbul Technical University Research Fund, ITU BAP, project code MGA-2020-42547. Finally, we would like to thank Lely Turkey Product Development for travel grant for Duygu Sesver.

{\small
\bibliographystyle{ieee_fullname}
\bibliography{annot.bib}
}

\end{document}